# Radiology Report-Conditional 3D CT Generation with Multi-Encoder Latent-diffusion Model

Sina Amirrajab[1], Zohaib Salahuddin[1], Sheng Kuang[1], Henry C. Woodruff[1,2], Philippe Lambin[1,2]

[1] The D-Lab, Department of Precision Medicine, GROW - Research Institute for Oncology and Reproduction, Maastricht University, 6220 MD Maastricht, The Netherlands

[2] Department of Radiology and Nuclear Medicine, GROW - Research Institute for Oncology and Reproduction, Maastricht University, Medical Center+, 6229 HX Maastricht, The Netherlands

## Abstract

**Background:**
Text-to-image latent diffusion models have recently advanced medical image synthesis, but applications to 3D CT generation remain limited. Existing approaches rely on simplified prompts, neglecting the rich semantic detail in full radiology reports, which reduces text–image alignment and clinical fidelity.

**Purpose:**
We propose Report2CT, a radiology report–conditional latent diffusion framework for synthesizing 3D chest CT volumes directly from free-text radiology reports, incorporating both *findings* and *impression* sections using multiple text encoder.

**Materials and Methods:**
Report2CT integrates three pretrained medical text encoders (BiomedVLP-CXR-BERT, MedEmbed, and ClinicalBERT) to capture nuanced clinical context. Radiology reports and voxel-spacing information condition a 3D latent diffusion model trained on 20,000 CT volumes from the CT-RATE dataset. Model performance was evaluated using Fréchet Inception Distance (FID) for real-synthetic distributional similarity and CLIP-based metrics for semantic alignment, with additional qualitative and quantitative comparisons against GenerateCT model.

**Results:**
Report2CT[1] generated anatomically consistent CT volumes with excellent visual quality and text–image alignment. Multi-encoder conditioning improved CLIP scores, indicating stronger preservation of fine-grained clinical details in the free-text radiology reports. Classifier-free guidance further enhanced alignment with only a minor trade-off in FID. We ranked 1st in the VLM3D Challenge at MICCAI 2025 on Text-Conditional CT Generation and achieved state-of-the-art performance across all evaluation metrics[2].

**Conclusion:**
By leveraging complete radiology reports and multi-encoder text conditioning, Report2CT advances state-of-the-art 3D CT synthesis, producing clinically faithful and high-quality synthetic data. These findings highlight the potential of report-driven generation for data augmentation, fairness, and robustness in medical imaging AI.

---

[1] Code will be available: https://github.com/sinaamirrajab/report2ct
[2] https://ctgen.vlm3dchallenge.com/

# Introduction

Text-to-image generation, the process of synthesizing realistic images from textual descriptions, has gained significant traction with the advent of text-conditional latent diffusion models (LDMs)[1]. These models generate images by iteratively denoising samples in a compressed latent space, enabling controllable synthesis guided by natural language prompts. While initially developed for natural images, their application to healthcare is rapidly expanding [2,3].

In medical imaging, text-conditional synthesis offers promising solutions to long-standing challenges, particularly data scarcity, orphan disease and privacy concerns[2,4]. High-quality synthetic data can enable data sharing while protecting patient confidentiality, can be used for virtual clinical trials[5,6], training of radiologists[7], augmenting training datasets for AI-based diagnostic[8] and segmentation[9,10] models, promote fairness and reduce bias[11], generate counterfactual cases for explainable AI[12], and simulate rare scenarios for robustness testing[13,14]. Recent works such as MediSyn[4], Text2CT[15], GenerateCT[16] demonstrate how synthetic medical images generated from text can improve downstream tasks.

Despite these advances, extending text-to-image diffusion to volumetric 3D scans (e.g., CT, MRI) remains highly challenging. Unlike 2D images, 3D medical volumes pose difficulties due to their high dimensionality, the need for strict inter-slice anatomical consistency, and the limited availability of paired text–image datasets. Several recent efforts have tried to address these barriers.

# Related work

### 2D Text-to-Image Generation in Radiology

Early work on text-conditioned diffusion in medical imaging was led by RoentGen[17] model, which fine-tuned a pretrained latent diffusion model on hundreds of thousands of chest X-rays paired with radiology reports. This adaptation to the medical domain enabled control over text-specified pathologies, and produced images that radiologists judged as realistic and semantically aligned. Synthetic data augmentation with RoentGen improved a disease classifier's accuracy. Chest-Diffusion[18] advanced efficiency and alignment by introducing a domain-specific CLIP text encoder trained on biomedical image–text pairs and a compact Transformer U-Net operating in the latent space. They achieved higher realism and superior pathology-specific alignment scores, demonstrating that domain-tailored architectures can deliver both fidelity and computational efficiency.

### 3D Text-Conditional Generation: MRI and CT

Initial efforts in 3D medical image synthesis focused on unconditional generation. Khader et al.[2] demonstrated that careful noise scheduling in a denoising diffusion model could yield realistic 3D MRI and CT volumes with strong slice-to-slice consistency. Müller-Franzes et al.[19] introduced *Medfusion*, a class-conditional latent DDPM, and showed across multiple modalities that diffusion models outperformed GANs in both fidelity and diversity.

Building on these foundations, GenerateCT[16] model pioneered free-text–to–3D CT generation via a cascaded pipeline: (1) a 3D Vision Transformer encodes low-resolution CT volumes into tokens; (2) a text–image transformer aligns these tokens with prompt embeddings; and (3) a text-conditioned super-resolution diffusion model reconstructs high-resolution slices. This hybrid 3D+2D strategy achieved state-of-the-art fidelity and alignment, while synthetic data improved multi-label abnormality detection.

Subsequent approaches further enhanced anatomical fidelity. MedSyn[20] introduced a hierarchical diffusion model that jointly generates high-resolution CTs and segmentation masks of key structures such as airways and vessels. Conditioning on both text and anatomical maps enabled preservation of fine structures like fissures and vascular branching. Molino et al.[21] proposed an end-to-end 3D latent diffusion framework that integrates a dual-encoder 3D CLIP trained on the CT-RATE dataset to establish a modality-specific text–image embedding space. This design enabled high-resolution synthesis without multi-stage upscaling and improved diagnostic model AUCs when synthetic CTs were used for training. Most recently, Text2CT[15] extended this end-to-end approach by introducing a flexible prompt encoder for diverse free-text clinical descriptions. By avoiding slice-wise super-resolution, Text2CT eliminated inter-slice discontinuities and reduced artifacts, achieving superior anatomical fidelity compared with GenerateCT and MedSyn.

**Limitation and Open Challenge**

Despite these advances, prior studies have largely relied on simplified conditioning text. Most employ short prompts or templated phrases derived from radiology reports, often focusing on a few keywords (e.g., *"left lung mass"* or *"right pleural effusion"*). While this reduces complexity, it sacrifices the nuanced clinical context embedded in full reports (e.g., *"Acute dyspnoea; afebrile. Known COPD. Findings: Spiculated pulmonary nodule with an associated ground-glass component, located in the right upper lobe; approximately 1.5 cm in diameter"*). As a result, text–image alignment scores (commonly measured with CLIP score) remain modest, reflecting incomplete transfer of semantic detail into the generated imagery.

Notably, none of the reviewed methods utilize the complete *findings* and *impression* sections of radiology reports, which together provide rich narrative descriptions of anatomical observations, pathological features, and diagnostic interpretations. These sections often contain subtle descriptors that, if incorporated, could substantially improve both anatomical accuracy and text–image alignment. The omission of such detailed context results in under-conditioned generations in which clinically relevant but less common features may be missing or inaccurately depicted.

We hypothesize that incorporating the complete *findings* and *impression* sections of the radiology report as conditioning input will provide richer semantic signals to the diffusion model, thereby enabling more precise and clinically faithful CT image synthesis.

## Contributions

We propose Report2CT (Figure 1), a text-conditional latent diffusion framework for synthesizing full 3D chest CT volumes at clinically relevant resolution, conditioned directly on detailed radiology reports. Our contributions are threefold:

1. Image compression network – High-resolution 3D CT volumes are encoded into a compact latent representation, reducing spatial dimensions by a factor of 4 to enable memory-efficient diffusion training.

2. Multi-encoder text representation – Three pretrained medical text encoders process the *findings* and *impression* sections independently, producing semantic embeddings that enrich text conditioning.

3. Text-conditional latent diffusion model – Latent-space CT volumes are generated via cross-attention mechanisms conditioned jointly on text embeddings and voxel-spacing information.

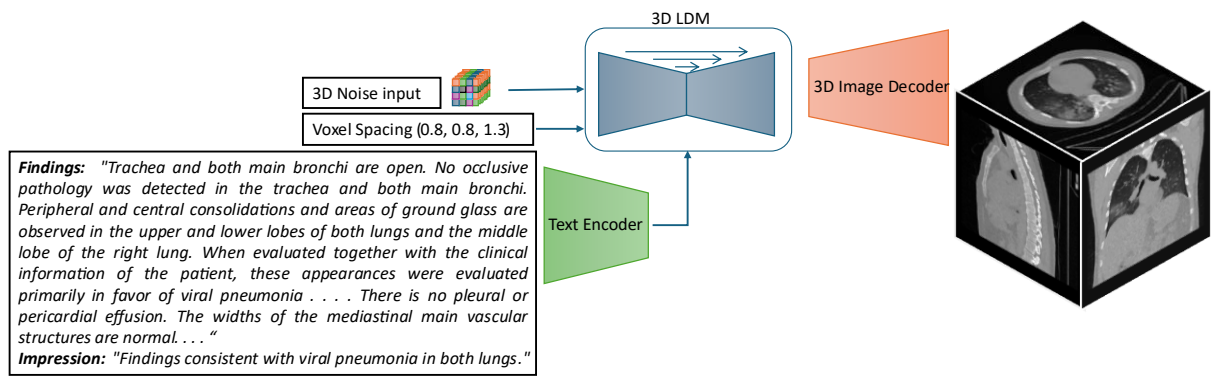

*Figure 1 Overview of the Report2CT framework, which generates 3D CT volumes from radiology reports. The findings and impression sections are encoded with medical language models and, together with voxel spacing, condition a 3D latent diffusion model to produce anatomically consistent and semantically aligned CT images.*

## Method

### Compression network

For image compression, we adopt the MAISI[22] network, pretrained on 39,206 3D CT volumes, to effectively map 480×480×256 voxel inputs into a latent space of size 4×120×120×64, enabling efficient training of the latent diffusion model without sacrificing anatomical detail. The latent diffusion model is trained on 20,000 3D CT volumes from the CT-RATE[23] dataset, each paired with the corresponding radiology report.

**Text Conditioning**

Our Report2CT framework follows the text-conditioning principles of latent diffusion models (e.g., Stable Diffusion[1]), extending the 3D U-Net denoiser with cross-attention mechanisms that allow the network to "attend" to the radiology report at every denoising step.

**Multi-Encoder Text Representation**

Radiology reports are typically structured into two main sections. The *findings* section documents the radiologist's direct observations from the images, covering both normal anatomy and any detected abnormalities, often with specifications on size, location, and morphological features. The *impression* section distills these observations into an overall clinical interpretation, highlighting the most relevant abnormalities and framing them in the context of the patient's condition.

To capture nuanced clinical information, we propose to process the *findings* and *impression* sections of each radiology report separately using three distinct pretrained medical text encoders, namely BiomedVLP-CXR-BERT[24], MedEmbed[25], and ClinicalBERT[26]. Each encoder produces a contextual embedding sequence from the tokenized input. We apply masked mean pooling to each sequence, concatenate the pooled vectors from all three encoders, and obtain a 2,560-dimensional embedding per section. The embeddings for *findings* and *impression* are concatenated channel-wise to form the final text-conditioning tensor. This multi-encoder setup is designed to capture nuanced clinical language and ensure that subtle descriptive cues are preserved.

**Conditioning on Voxel Spacing**

In addition to text, the model receives a voxel-spacing embedding, which is concatenated to the text embeddings before projection. This ensures that generated CT volumes not only match the described findings but also adhere to the desired resolution and anisotropy.

**Classifier-Free Guidance**

Classifier-Free Guidance (CFG)[27] is a technique in text-conditional latent diffusion models that enables a controllable balance between fidelity to a text prompt and sample diversity. To enable CFG, the text conditioning vector is replaced with zero vector with a certain probability, making the LDM to learn unconditional generation. During inference, the model generates two predictions per timestep: one conditioned on the text prompt ($\epsilon_{cond}$) and one without any conditioning ($\epsilon_{uncond}$). These are combined into the final noise estimate as:

$$\epsilon = \epsilon_{uncond} + s \cdot (\epsilon_{cond} - \epsilon_{uncond})$$

Where $s$ is the CFG scale factor that amplifies the influence of the prompt on generation. Lower scale values yield more diverse outputs with weaker prompt alignment, while moderate scales (e.g. 3–7) typically offer a reliable trade-off between semantic coherence and variability. Higher CFG scales may produce outputs with visually precise alignment to the text prompt but at the cost of reduced diversity, and potentially degraded quality if the scale is set excessively high and can even result in over-specified or hallucinated content when the model overfits the textual input.

Figure 2 illustrates how text encoders, voxel spacing, and noise scheduling interact in the latent diffusion process.

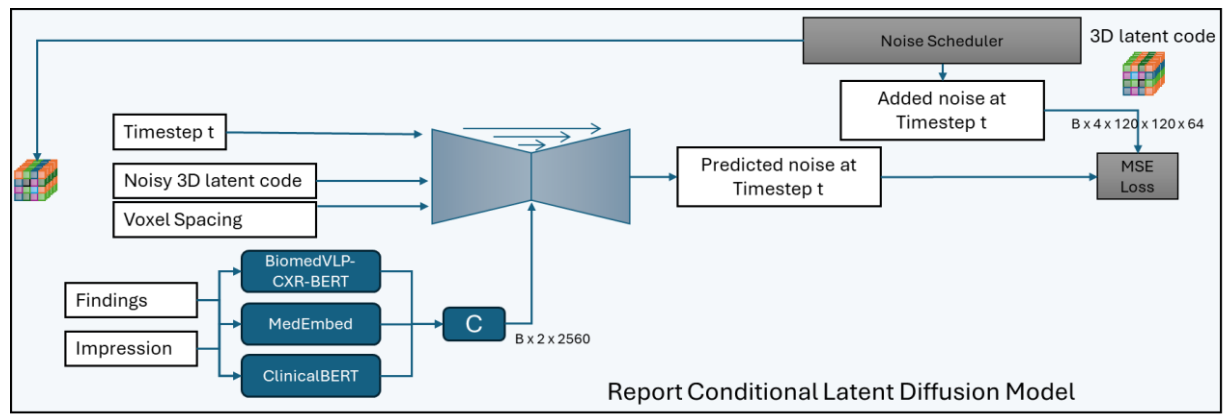

*Figure 2 Architecture of the proposed Report2CT model. The findings and Impression sections are encoded separately using three pretrained medical text encoders (BiomedVLP-CXR-BERT, MedEmbed, and ClinicalBERT). Their pooled embeddings are concatenated into a 2,560-dimensional vector, which is combined with voxel-spacing information and injected into the latent diffusion model via cross-attention modules. The model learns to predict noise at each timestep, guided by a noise scheduler, and reconstructs 3D CT volumes from compressed latent codes.*

**Evaluation metrics**

To evaluate how well synthetic data approximates real data in terms of quality and diversity, we employed the Fréchet Inception Distance (FID)[28] with a 2.5D feature extraction strategy. Each CT volume was preprocessed (reorientation, optional resampling, padding/cropping, intensity clipping) and sliced along the three orthogonal planes (XY, YZ, ZX). Single-channel inputs were replicated to three channels, normalized, and passed through a fixed RadImageNet[29] ResNet-50 to extract 2D features, which were spatially averaged per slice and aggregated across cases. For each plane, empirical feature distributions were obtained for real and synthetic images, and FID was computed as the Fréchet distance between their Gaussian fits. The final score was the average of the three plane-wise FIDs, with lower values indicating greater similarity between synthetic and real images.

We additionally assessed image quality and semantic consistency using a CT-adapted CLIP model.[23]. Each generated CT volume was resampled and cropped to a fixed resolution, paired with its corresponding radiology text prompt and reference CT, and encoded into a shared image–text latent space. Following GenerateCT model, two metrics were computed:

CLIPScore-T2I: cosine similarity between the generated image and its conditioning prompt.

CLIPScore-I2I: cosine similarity between the generated image and its paired real CT.

Higher CLIP scores reflect improved alignment with the clinical description (T2I) and greater fidelity to the ground-truth anatomy (I2I). Taken together, FID captures global distributional similarity between real and synthetic images, while CLIP-based scores directly evaluate semantic faithfulness and image–text alignment. This complementary evaluation provides a more comprehensive assessment of synthetic CT data quality.

**Implementation details**

All CT scans, originally acquired at varying resolutions and matrix sizes, were resampled to a uniform volume of 480 × 480 × 256 voxels. Intensities were clipped to the clinical range of [−1000, 1000] Hounsfield units and subsequently min–max normalized to [0, 1]. As in the MAISI framework, resampling alters the voxel spacing, which we explicitly provide as an additional conditioning input to the diffusion model. The implementation was based on MONAI[30] and its generative AI extension[31].

For latent representation learning, we employed the pre-trained AutoencoderKL-MAISI network to compress images into a latent space of size 4 × 120 × 120 × 64. The compressed latents were then used as inputs for a 3D diffusion U-Net with hierarchical channel widths [64, 128, 256, 512], multi-head cross-attention in the last two resolution levels, and cross-attention dimension 2560 to accommodate multi-text embeddings. Both findings and impression sections of the radiology reports were encoded independently into contextual embeddings (concatenated at training time) and served as conditioning signals for the diffusion model.

For text conditioning, we integrate multiple pre-trained clinical language models, including MedEmbed-large-v0.1, ClinicalBERT, and BiomedVLP-CXR-BERT-specialized. Each report section (findings and impression) was independently tokenized and encoded by all three models. Mean pooling (weighted by the attention mask) was applied to the final hidden states of each encoder to obtain dense representations. These were concatenated into a 2560-dimensional embedding (768 + 768 + 1024). The concatenated pooled embedding was used as the sole text-conditioning signal for cross-attention in the diffusion U-Net. This concatenation strategy was chosen to leverage complementary strengths from general-domain, clinical-domain, and radiology-specialized text encoders, thereby enriching the semantic representation of radiology reports for conditioning.

Training was conducted on two NVIDIA H100 NVL GPUs using distributed data parallelism with mixed precision. The model was trained with a batch size of 2 and a learning rate of $1 \times 10^{-4}$, optimized with Adam and a polynomial decay scheduler. The diffusion process followed a Rectified Flow (RFlow)[32] scheduler with 1,000 steps and a scale factor of ~1.4. To support classifier-free guidance, conditioning embeddings randomly dropped with a probability of 0.15 during training, improving robustness and enabling guided sampling at inference.

## Experiments and Results

Figure 3 presents a qualitative comparison between the proposed Report2CT model and GenerateCT model, showing three orthogonal views alongside the corresponding text prompts. Report2CT produced 3D volumes with greater anatomical consistency and higher visual quality, indicating the effectiveness of conditioning on full radiology reports.

**The effects of CFG scale**

We investigated the effect of different CFG scales on the synthetic image quality and image-text alignment. Figure 4 shows the impact of varying CFG scales on image synthesis quality and the degree of alignment with the input text. We can observe that varying these scales leads to differences in both the visual quality and how closely the generated images reflect the textual descriptions.

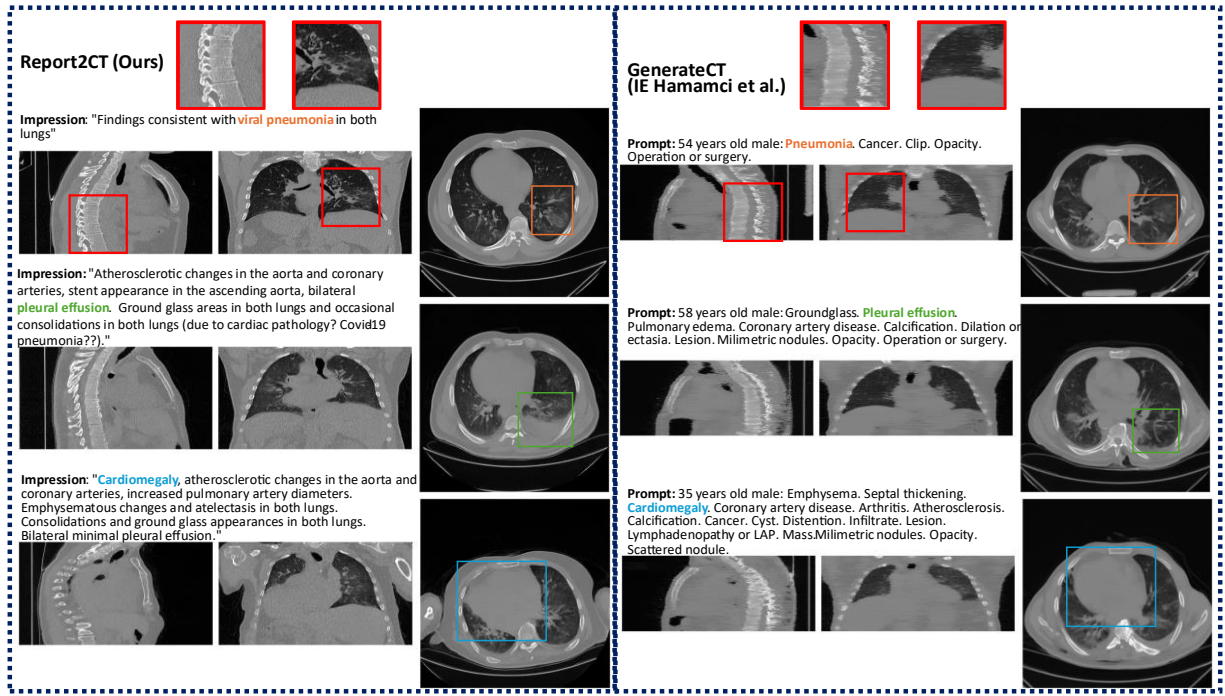

*Figure 3 Qualitative comparison between our proposed Report2CT model and the GenerateCT model for different abnormalities (lung pneumonia in orange, pleural effusion in green, and cardiomegaly in blue). The red zoomed-in region highlights the consistency of Report2CT across orthogonal views. Both findings and impressions were used for synthesis; for brevity, only impression texts are shown.*

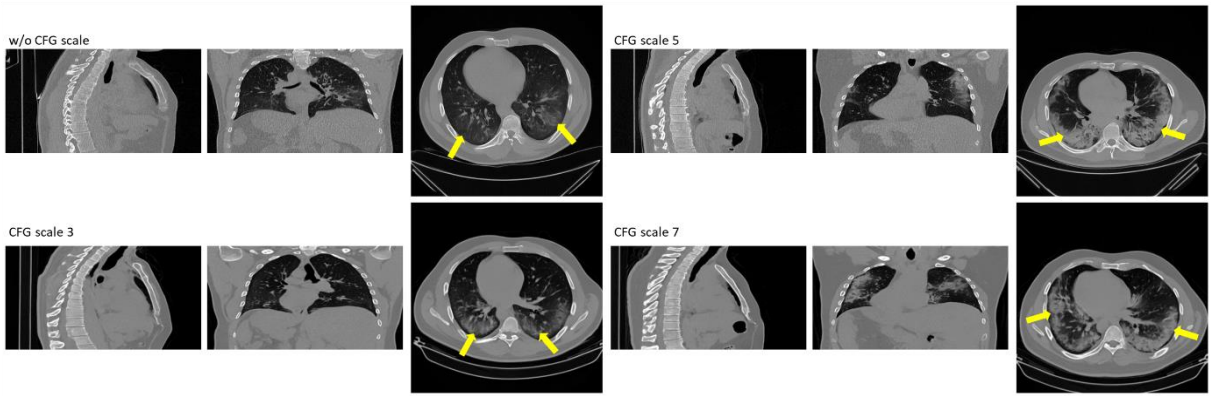

*Figure 4 Impact of varying CFG scales on image synthesis quality and the degree of alignment with the input text.*

**The effects of multiple text encoders**

We evaluated the impact of using multiple text encoders in combination with different CFG scales. As shown in Figure 5, incorporating three encoders markedly improved the preservation of fine-grained details from the *findings* and *impression* sections, achieving higher CLIP scores compared to single-encoder conditioning. This demonstrates that multi-encoder representations enhance semantic alignment between text and generated images. Consistent with the visual results in Figure 4, variations in CFG scale further influenced both visual quality and text–image coherence. These trends were reflected in our evaluation metrics: improvements in CLIP scores confirmed stronger text–image alignment, while changes in FID (Figure 6) quantified the trade-off between alignment and overall image quality. Together, these findings underscore the importance of optimizing both text encoding and guidance strength to generate high-quality, clinically aligned synthetic CT volumes.

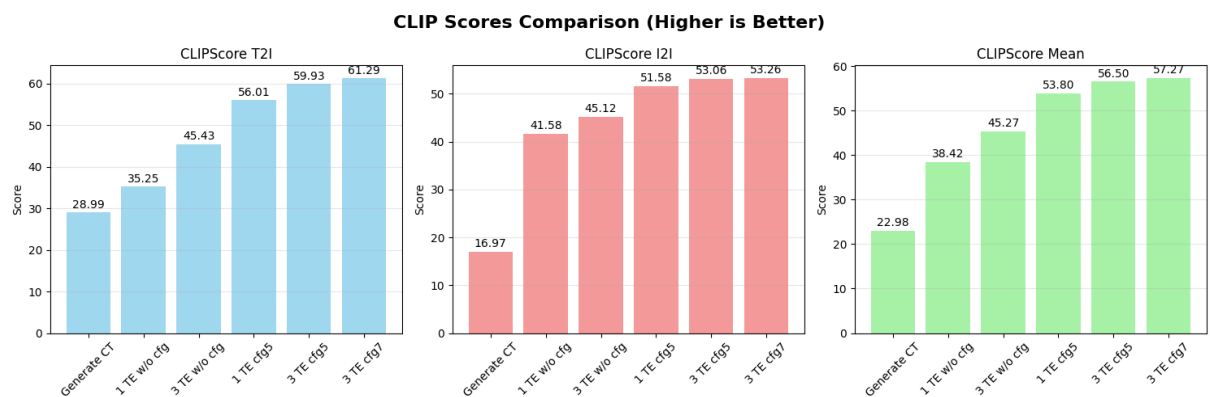

*Figure 5 CLIP scores (higher is better) comparison between Generate CT model and the proposed Report2CT models with one text encoding (1TE) and three text-encoding (3TE) with and without classifier free guidance (cfg) scales.*

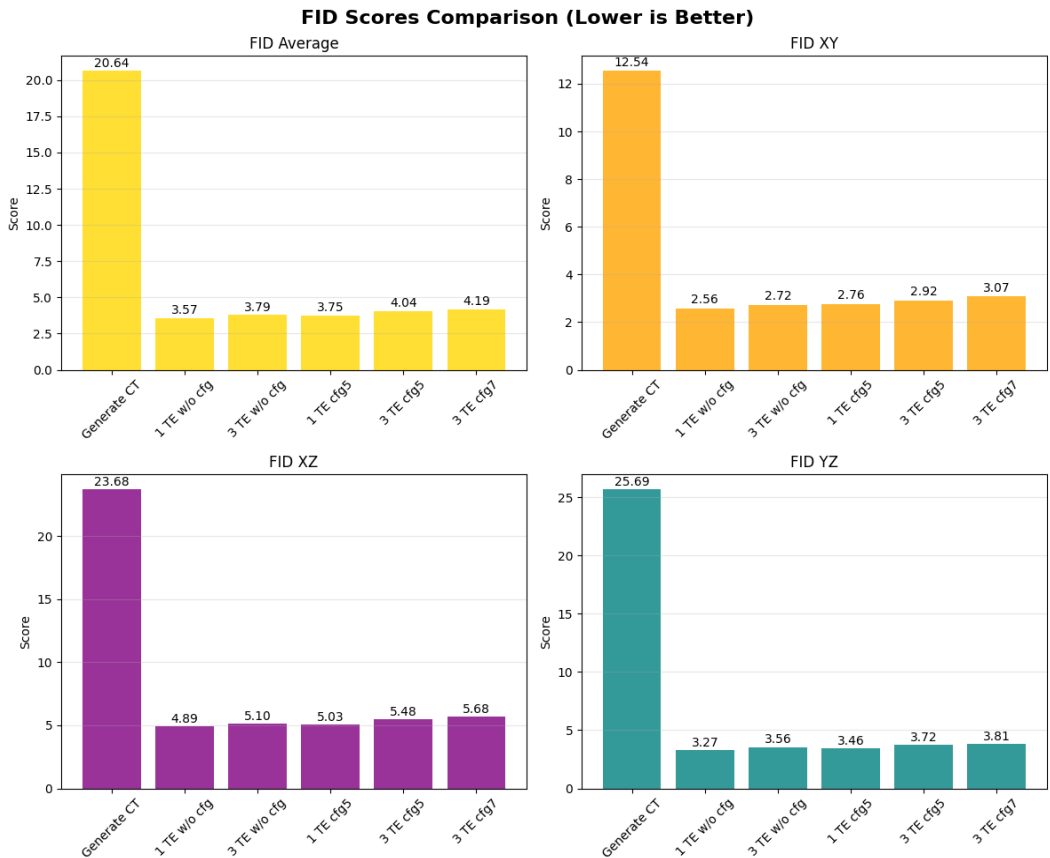

*Figure 6 FID scores (lower is better) for comparing the quality of the generated images using Generate CT model and the proposed Report2CT models with one text encoding (1TE) and three text-encoding (3TE) with and without classifier free guidance (cfg) scales*

## Discussion

In this paper, we introduced Report2CT, a radiology report–conditional diffusion model for generating 3D CT images from full radiology reports, including both the *findings* and *impression* sections. By incorporating three pretrained medical text encoders, the model achieved state-of-the-art text–image alignment, as reflected in higher CLIP scores. Adjusting the classifier-free guidance scale further enhanced alignment, with only a minor trade-off in FID, suggesting a small reduction in visual quality. Across all evaluation metrics, Report2CT consistently outperformed GenerateCT by a substantial margin. Report2CT also ranked 1st in Task 4 of the VLM3D Challenge at MICCAI 2025 on Text-Conditional CT Generation[3], demonstrating its robustness and generalizability in a competitive international benchmark. These results highlight that leveraging the full radiology report text together with multiple pretrained text encoders was critical to achieving consistent improvements across all evaluation metrics.

**Limitations and future work**

A primary limitation of Report2CT is its computational demand. Training on two NVIDIA H100 NVL GPUs (94 GB each) required approximately one hour per epoch, restricting the number of experiments and limiting applicability in resource-constrained settings such as hospitals. In contrast, inference is less demanding and can be performed on GPUs with smaller memory (e.g., 40 GB). Another limitation is the reliance on the text from both the *findings* and *impression* sections for conditioning, which may not always be available and typically require expert radiologists to produce comprehensive reports. One potential solution is to leverage AI-based report generation models to create detailed synthetic reports for open-source CT datasets, which could then be used for image synthesis. Direct comparison with Text2CT was not feasible due to the unavailability of publicly released code or synthetic data paired with input text.

---

[3] https://ctgen.vlm3dchallenge.com/

Future research will explore combining CT image generation with report generation systems and large language models (LLMs) to enable fully automated creation of clinically plausible reports for conditioning. We also plan to assess the utility of synthetic CTs in improving abnormality detection, particularly for rare diseases and edge cases by selectively synthesizing such examples. Further work will focus on reducing computational requirements, expanding model generalizability across datasets and modalities, and investigating clinical applications of synthetic CT data for diagnostic model training and/or validation of AI models, training of radiologists[7], and In Silico clinical trials[5].

## Conclusion

We presented Report2CT, a latent diffusion framework for 3D chest CT generation conditioned on complete radiology reports. By integrating multiple pretrained medical text encoders and leveraging both the *findings* and *impression* sections, Report2CT achieved superior text–image alignment and higher fidelity than existing models. Experimental results demonstrated that our model produces anatomically consistent volumes and better preserves clinically relevant details, as reflected in CLIP and FID metrics. These findings highlight the potential of report-driven generative models to advance data augmentation, explainability, and robustness in medical imaging AI.

## Grants and Founding

Authors acknowledge financial support from the European Union's Horizon research and innovation programme under grant agreement: CHAIMELEON n° 952172 , ImmunoSABR n° 733008, EuCanImage n° 952103, IMI-OPTIMA n° 101034347, AIDAVA (HORIZON-HLTH-2021-TOOL-06) n°101057062, REALM (HORIZON-HLTH-2022-TOOL-11) n° 101095435, RADIOVAL (HORIZON-HLTH-2021-DISEASE-04-04) n°101057699, GLIOMATCH n° 101136670 and EUCAIM (DIGITAL-2022-CLOUD-AI-02) n°101100633. The research of H.C.W. and S.K. is partially supported by the Dutch Cancer Society (KWF Kankerbestrijding) (project no. 2021-PoC/14449).